\documentclass[letterpaper, 10 pt, conference]{ieeeconf}
\IEEEoverridecommandlockouts                            
\usepackage[T1]{fontenc} 
\usepackage{enumerate}
\usepackage{amsfonts}
\usepackage[colorinlistoftodos]{todonotes}
\usepackage[font=small,skip=12pt]{caption}
\usepackage{float}
\usepackage{subcaption}
\usepackage[ruled,linesnumbered]{algorithm2e}

\usepackage{algorithmicx}
\usepackage{booktabs}
\makeatletter
\let\NAT@parse\undefined
\makeatother
\usepackage{hyperref}
\usepackage{verbatim}
\usepackage{comment}
\usepackage{amsmath} 

\algnewcommand{\LeftComment}[1]{\Statex \(\triangleright\) #1}

\title{\LARGE \bf
\textit{Real-to-Sim} Registration of Deformable Soft Tissue with Position-Based Dynamics for Surgical Robot Autonomy}

\author{Fei Liu$^{\dagger, 1}$ \IEEEmembership{Member, IEEE}, Zihan Li$^{\dagger, 1}$, Yunhai Han$^{1}$, Jingpei Lu$^{1}$, Florian Richter$^{1}$ \IEEEmembership{Student Member, IEEE}\\ and Michael C. Yip$^1$ \IEEEmembership{Senior Member, IEEE}
\thanks{$\dagger$ Equal contributions.}
\thanks{$^1$Advanced Robotics and Controls Lab, University of California San Diego, La Jolla, CA 92093 USA. {\tt\small \{f4liu, zil027, y8han, jil360, frichter, yip\}@ucsd.edu}}%}%
}

\begin{document}

\maketitle 
\thispagestyle{empty}
\pagestyle{empty}

\begin{abstract}
Autonomy in robotic surgery is very challenging in unstructured environments, especially when interacting with deformable soft tissues.
% , as tissues tend to deform significantly during manipulation. 
% This creates a challenge for model-based control methods that must account for deformation dynamics during tissue manipulation.
The main difficulty is to generate model-based control methods that account for deformation dynamics during tissue manipulation.
% compared to traditional control and task automation which have been successfully demonstrated in a variety of well-structured environments.
% Taking advantage of vision-based perception, one can capture the geometric changes within the scene. 
% However, the dynamic properties of tissue cannot be captured and predicted with only visual information. 
Previous works in vision-based perception can capture the geometric changes within the scene, 
% however, integration with dynamic properties to achieve accurate and safe model-based controllers has not been considered before.
however, model-based controllers integrated with dynamic properties, a more accurate and safe approach, has not been studied before.
% Because manipulating deformable objects results in a coupling of mechanics between the robot and the environment, 
Considering the mechanic coupling between the robot and the environment,
it is crucial to develop a registered, simulated dynamical model. 
% for model-based methods to incorporate the mechanics of a 
% in order to achieve accurate and safe control. 
% This assumes though that the simulation is accruately modeling the environment, and often prior information about stiffness and constraints of tissues are only known to a first order. 
In this work, we propose an online, continuous, \textit{real-to-sim} registration method to bridge 3D visual perception with position-based dynamics (PBD) modeling of tissues. The PBD method is employed to simulate soft tissue dynamics as well as rigid tool interactions for model-based control. Meanwhile, a vision-based strategy is used to generate 3D reconstructed point cloud surfaces based on real-world manipulation, so as to register and update the simulation.
% resulting in this real-to-sim translation. 
% The registration algorithm is applied as an additional dynamic constraint for PBD that corrects the surface meshes but still maintains the inner structures of the simulated soft tissue. 
To verify this real-to-sim approach, tissue experiments have been conducted on the da Vinci Research Kit. Our real-to-sim approach successfully reduces registration error online, which is especially important for safety during autonomous control. Moreover, it achieves higher accuracy in occluded areas than fusion-based reconstruction.

% the da Vinci Surgical robot with an end-effector gripping the target objects along the regulated trajectories. 

\end{abstract}
 
\section{Introduction}
% pipeline:\\
% 1. irreplaceable impact of vision-based correction in deformation simulation. may also point out its potential in control\\
% 2. define the word 'registration' mentioned in this paper (it's actually a vision-based correction)\\
% 3. related work: shape matching, Position-based dynamics, sparse surface constraint (Weiss's paper)\\

% Surgical robotic autonomy has been a spotlighted field in recent years, both in industry and academic \cite{Yip2017Automation}.
% It helps to increase accuracy in sub-task automation
Surgical robotic autonomy has drawn significant interest in recent years, as it may help ease surgeon fatique, reduce human errors, or address lack-of-access to timely, life-saving surgery in remote or under-served communities \cite{Yip2017Automation}. Regardless of what type of surgical task is being performed,
% The research interest in sub-task automation has been raised in literature,
% such as tissue retraction \cite{Attanasio_2020}, suturing \cite{Pedram_2017} and cutting \cite{Murali_2015}. In these works, the levels of autonomy \cite{Yang_2017_ScieceRobotics} is restricted within guidance or conditional level.
% 3D reconstruction and tracking of the surgical scene is required -- it provides the 3D geometric and mechanical information necessary to establish the underlying tissue models used during control synthesis, which are tasked with safely manipulating tissues.
which is essential for manipulating tissues safely.
%real-time understanding and representation of the unstructured intraoperative surgical environment, particularly in interacting with deformable soft tissues.

Different approaches to 3D reconstruction and tracking from cameras have been proposed for dynamic
and deformable environments, such as structure from motion (SFM) \cite{Cheema_2019}, simultaneous localization and mapping (SLAM) \cite{Mahmoud_2019, Song_2018_SLAM}, and fusion-based model-free
tracking \cite{SuperLi2019, lu2020super}. A more comprehensive review can be found in \cite{SuperLi2019}.
%Visual perception can capture the non-rigid deformation by providing real-time reconstruction and tracking of soft tissue.
However, visual information alone is not capable of providing internal dynamical properties of soft tissues, such as mechanics, inertial properties -- the features that are needed for accurate model-based control.
% kinetic motion, elastic deformation.
% It restricts the autonomous tasks for traditional model-based control methods
%The lack of dynamics information  the autonomous tasks using traditional model-based control methods, for example, model predictive control (MPC) of surgical manipulation, which requires prediction of tissue behavior under loads in order to properly assess optimal trajectories and optimize control actions.

\begin{figure}[t!]
\setlength{\belowcaptionskip}{-0.3cm} 
\vspace{2mm}
\begin{subfigure}{0.24\textwidth}
\includegraphics[width=1\textwidth,height=1.1in]{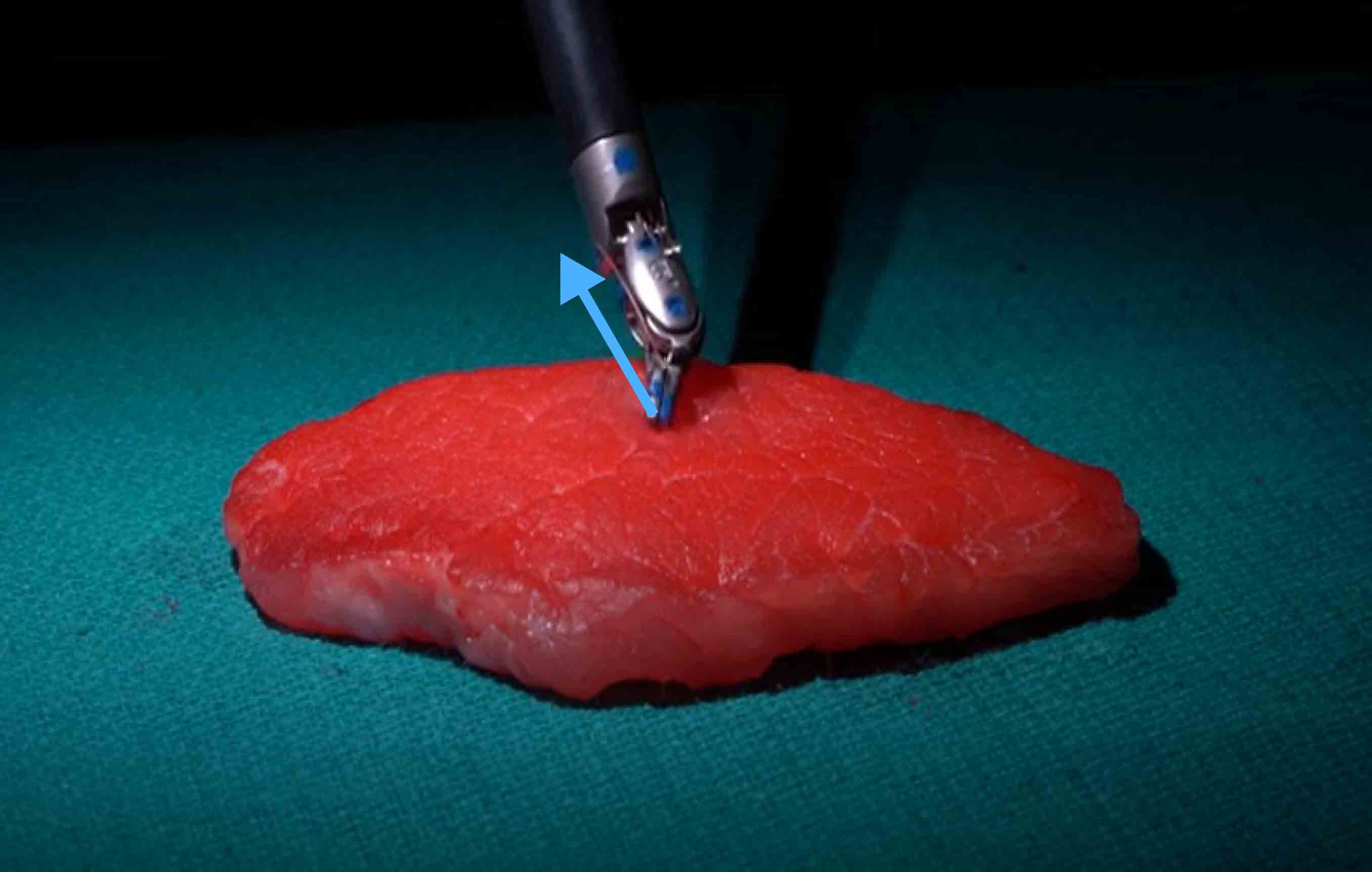}
\vspace{-0.12in}
\end{subfigure}
\begin{subfigure}{0.24\textwidth}
\includegraphics[width=1\textwidth,height=1.1in]{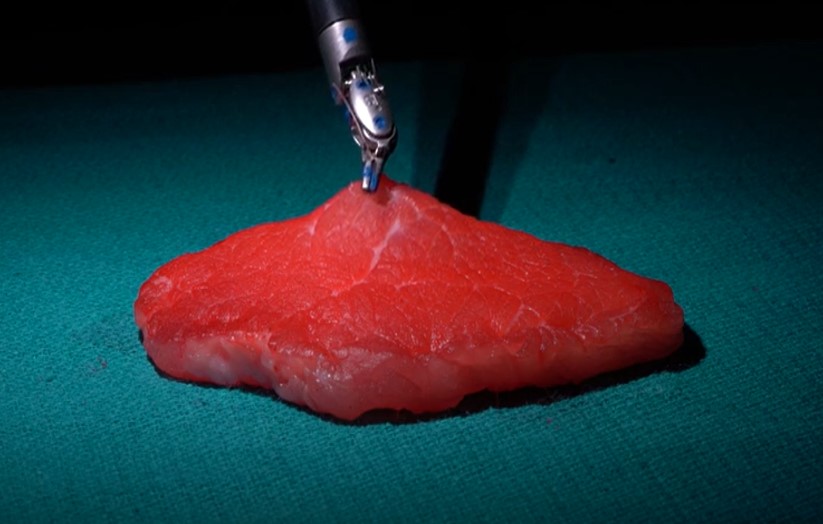}
\vspace{-0.12in}
\end{subfigure}
\begin{subfigure}{0.24\textwidth}
\includegraphics[width=1\textwidth]{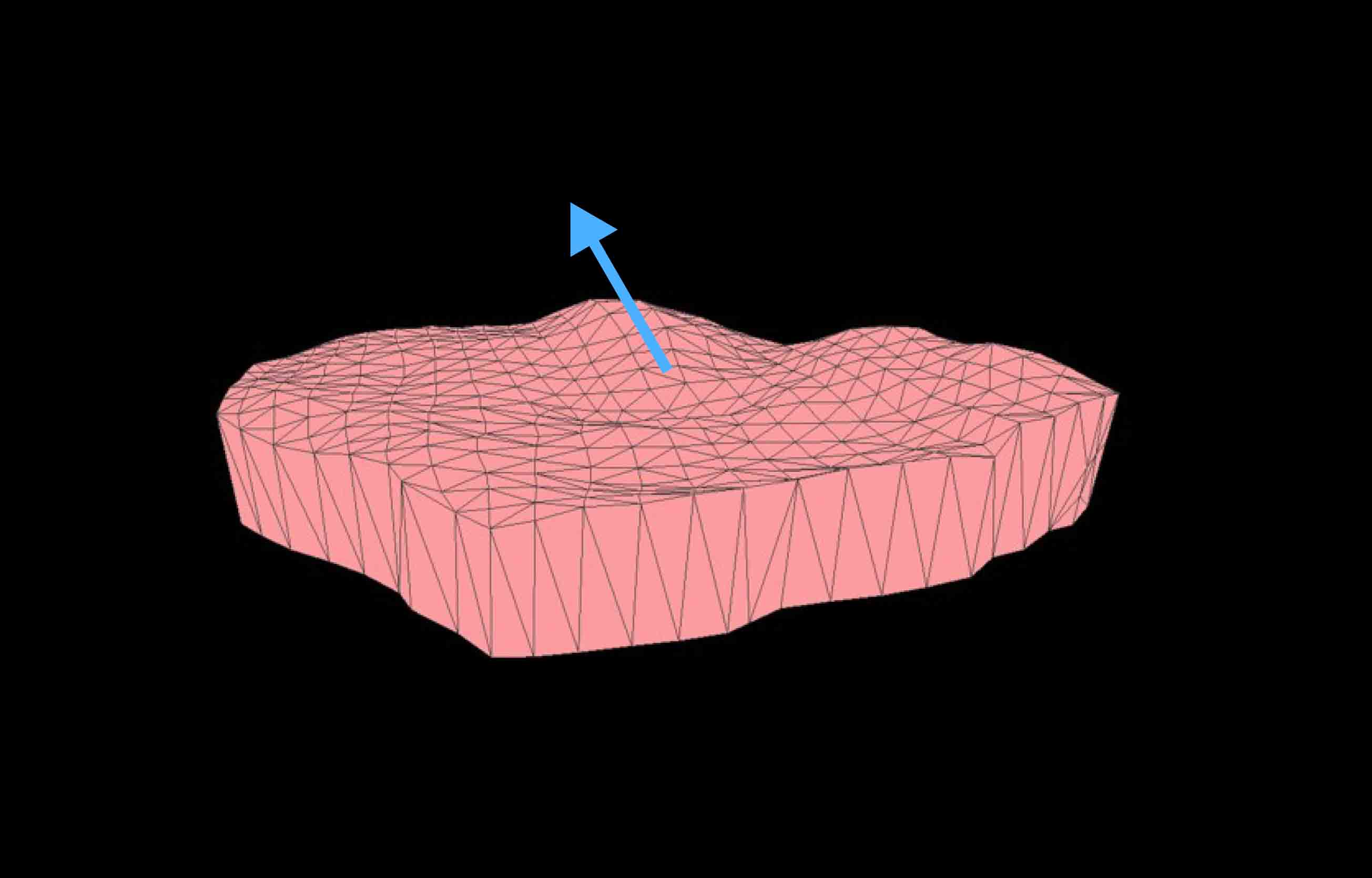}
\end{subfigure}
\begin{subfigure}{0.24\textwidth}
\includegraphics[width=1\textwidth]{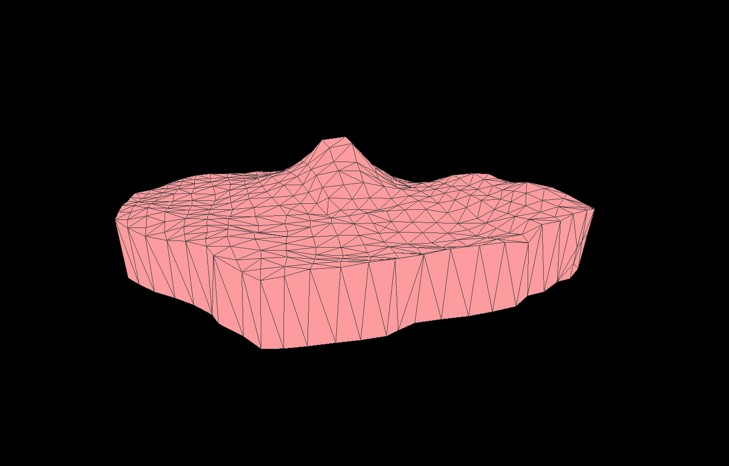}
\end{subfigure}
\caption{A demonstration of real-to-sim registration for deformable tissue. The top figures show the real images, and the bottom figures show the corresponding registered meshes before (left) and after (right) manipulation using our method. The blue arrow indicates the grasping direction.}
\label{fig:cover}
\vspace{-0.1in}
\end{figure}

To address the aforementioned problem, several works have been conducted to estimate the partial tissue dynamics and deformations by using reinforcement learning \cite{Shin_2019}, deformation Jacobian \cite{Alambeigi_2018}, and adaptive estimator \cite{Zhong_2019}. However, these methods require the complete observation of the deformable tissue and are not able to handle the occlusions.
A Finite Element Method (FEM) implemented with the SOFA framework \cite{Faure_2012_SOFA} was used in  \cite{Adagolodjo_2019_TRO}  during  procedures involving inserting needles into tissues. However, FEM as a general strategy has a significant problem that one cannot explicitly apply position constraints on the simulation easily, so the registration between real world and simulation cannot be well defined. 
Works in autonomous debridement \cite{Aghajani_2019} and tissue tensioning \cite{Thananjeyan_2017_ICRA, Nguyen_2019} applied learning method to identifying proper tissue properties from visual input.
However, none of these works considered the physical dynamics explicitly. 
They directly extract control policies from vision rather than establish an underlying model and solve the model-based control problem, which limits the performance beyond the training environments. 
%Furthermore, only 2D deformations were considered. %Also, none of them can handle occlusion.
% It has been studied in the past decades, which can be divided into mesh-based (e.g., FEM, finite element method \cite{Faure_2012_SOFA, Tang_2014}) and mesh-free methods (e.g., MSM, mass-spring system modeling \cite{Duan_2016, Wang_2014}). It is a trade-off between accuracy and real-time performance due to high computational complexity. 

Another way to integrate tissue dynamics is to build a physical-based surgical simulator.
In computer graphics, position-based dynamics (PBD) is a popular method of simulating object deformation \cite{Macklin_2017} in real-time. It has shown great potential in the application for surgical simulation scenarios, such as biopsies \cite{Tagliabue_2019} and cutting \cite{Berndt_2017}.

Taking advantage of fast real-time PBD simulation, we intend to bridge the gap between visual perception and physical tissue dynamics modeling through an online, continuous, real-time registration method. We call this \textit{real-to-sim} registration. The method incorporates the point cloud extracted from cameras at each frame to update the simulated surface particles in PBD. This points-to-particles correspondence can be viewed as a surface constraint and solved as a registration cost function by gradient descent.
%It can also natively link the fusion data (i.e., surface point cloud ) with the particle representation in PBD. 
To the best of our knowledge, this is the first work to perform deformation registration for simulation using PBD,
% simulated position-based dynamics 
relying on real-world observation of tissue deformation (Fig. \ref{fig:cover}). %Meanwhile, non-rigid registration remains an open research problem in the general context of surgical robotics.
The contributions of our method are summarized as,
\begin{enumerate}
    \item a position based framework for surgical simulation involving physical constraints (i.e. distance, volume represented by particles, etc.),
    \item a real-to-sim matching algorithm for registration is applied as an additional dynamic constraint for PBD,
    \item integration of surgical perception framework (SuPer) proposed in \cite{SuperLi2019}, which can potentially improve the accuracy of fusion-based reconstruction in the occluded areas.
\end{enumerate}
Our method was implemented on a da Vinci Research Kit \cite{Kazanzides_2014_dvrk}. Multiple tissue manipulation experiments were conducted to highlight its effectiveness and accuracy. We believe that this \textit{real-to-sim} method is a fundamental step towards generalizable surgical automation.
\section{Simulation}
\label{sec:PBD_sim}

\subsection{Position-based Dynamics (PBD)}
Physical simulation has been studied in the past decades and can be classified into mesh-based (\cite{Faure_2012_SOFA, Tang_2014}) and mesh-free methods (\cite{Duan_2016, Wang_2014}). For all methods, there is an inherent trade-off between physical accuracy, computational stability, and real-time performance. Unlike physical simulators, the PBD method provides a  real-time solver and stable time integration scheme that makes it fast and robust to use in practice.
Different materials are identified not by their physical parameters but through constraint equations which define particle positions and position-derivatives.
This representation of positional evolution can naturally build the link from visual perception to image data, as it could force a topological constraint on the surface particles of a scene. 
It also allows us to combine different types of geometrical constraints (such as distance, volume, etc.). 
More details of PBD method can be found in \cite{Macklin_2017}. 
% We give some background on this method here, but refer to \cite{Macklin_2017} for more details. 
% We detail the proposed flowchart for our method in Fig. \ref{}.  

% including both vision-based constraints and physical-based constraints.
% Position-based dynamics method is an approach which ignores the velocity and acceleration layer and directly update the positions.

% \subsubsection{Simulation Process}
In our research, the simulated object is defined as a set of $N$ particles and $M$ constraints. Given the current position $\mathbf{x}$ and velocity $\mathbf{v}$ of the particles, the simulation process is described in Algorithm \ref{alg:PBD_simu}. A force $\mathbf{f}_{ext}$ acts on each particle, which only includes gravity in this work.
% This includes local particle attraction forces describing the stiffness of the tissue as well as gravity. 
%, and, the driving force exerted on grasping particles by the manipulator. However, in most cases, the driving force is unknown. 
When a local set of particles is grasped by a manipulator, their positions are constrained to the manipulator's trajectory under the assumption that they are fixed to the end effector. 
% \vspace{-0.3cm}
\begin{algorithm}[h]
\SetAlgoLined

$\mathbf{x}^* = \mathbf{x}^{t} + \Delta t \zeta \mathbf{v}^t + \Delta t^2 \mathbf{M}^{-1} \mathbf{f}_{ext}(\mathbf{x^t}) $ \Comment{{\scriptsize prediction}}
\\

% \For{particle $i \in N$ }{
% apply velocity damping $\mathbf{v}_{i} = \zeta \mathbf{v}_{i}$; \\
% apply forces $\mathbf{v}_{i} = \mathbf{v}_{i}+\Delta t \mathbf{f}_{ext}\left(\mathbf{x}_{i}\right)/m_{i}$; \\
% predict position $\mathbf{x}_{i}^{*} = \mathbf{x}_{i}+\Delta t \mathbf{v}_{i}$; \\
% $x_i^{t+1} = x_i^{*}$; \\
% } 

% \Comment{constraint solving step} \\
\While{ {\rm iter} $<$ SolverIterations}{

\For{constraint $C \in \mathcal{C}$}{
% \For{particle $i \in N$}{
% \text{Compute} 
% $\left[\Delta \mathbf{x}_i \right]_\text{c} \leftarrow \mathrm{solveConstraint}(\mathbf{x^{*}})$ \\
$\left[ \Delta \mathbf{x} \right]^{\text{iter}}_C = 
% \lambda_c \cdot
\nabla_{\mathbf{x}^{*}} C $  \Comment{{\scriptsize constraint solving}} \\
$\mathbf{x^{*}}=\mathbf{x^{*}} + \left[\Delta \mathbf{x} \right]^{\text{iter}}_C $
}

}
% \Comment{update step} \\
$\mathbf{x}^{t+1} = \mathbf{x^{*}}$ \Comment{{\scriptsize update position}} \\
$\mathbf{v}^{t+1} = \left(\mathbf{x}^{t+1}-\mathbf{x}^{t}\right)/\Delta t$ \Comment{{\scriptsize update velocity}} \\
% \For{particle $i \in N$}{
% update velocity $v_{i} = \frac{1}{\Delta t}\left(x_{i}^{t+1}-x_{i}^{t}\right)$
% }
\caption{Simulation Process}
\label{alg:PBD_simu}
\end{algorithm} 

% \subsubsection{The Gauss-Seidel Method}
% In Algorithm \ref{Alg_simu}, the position correction $\Delta \mathbf{x_i}$ for particle $i$ in each iteration can be computed through Gauss-Siedel Method \cite{Mull2017PBD}.

% The integration of real-to-sim registration in PBD method is the main contribution of this paper. Besides, there are other  that are used for the simulation.\\
% \vspace{-0.7cm}
\vspace{-0.15cm}
\subsection{Geometric Constraints}
We include several geometric constraints for simulation of the particle dynamics to generate the soft tissue deformation.
% \subsubsection{\textbf{Distance Constraint}} 
% The mass-spring model (MSM) has introduced into the surgery simulation of deformable objects several years ago \cite{Meier2005MassS}. It is reasonable to represent the particle's connection as mass-spring networks to capture the dynamics. Consider the distance constraint function between two particles $\mathbf{x_1}$, $\mathbf{x_2}$, 
% \begin{equation}
%     C_{dis} \left(\mathbf{x_{1}}, \mathbf{x_{2}}\right)=\left|\mathbf{x_{1,2}}\right|-\mathbf{d_0}.
% \end{equation}
% where $\mathbf{d_0}$ is the initial distance between $\mathbf{x_1}$ and $\mathbf{x_2}$. The formulas of solving the position corrections $\Delta \mathbf{x}$ can be found in \cite{Macklin_2017}.

% \subsubsection{\textbf{Volume Preservation}}
% Volume preservation constraint is important for the simulation of incompressibility, which is one of the main mechanical behaviors of deformable objects \cite{Duan2014Volume}. For tetrahedral meshes, such a constraint has the following form to preserve the volume of a single tetrahedron:
% \begin{equation}
% C_{vol}\left(\mathbf{x_{1}, x_{2}, x_{3}, x_{4}}\right)=\frac{1}{6}\left(\mathbf{x_{2,1}} \times \mathbf{x_{3,1}}\right) \cdot \mathbf{x_{4,1}} - v_0
% \end{equation}
% where $\mathbf{x_1,x_2,x_3,x_4}$ are the four corner particles of the tetrahedron and $v_0$ is its initial volume. The position correction can be obtained using a similar formula as distance constraint.
\subsubsection{\textbf{Distance Constraint} \& \textbf{Volume Preservation}} 
In \cite{Han2020Simu}, the authors proposed a 2D PBD-based surgical simulation framework. In this work, we adapt the same distance constraint and replace the triangle area preservation with tetrahedron volume preservation. 

\begin{equation}
\setlength{\abovedisplayskip}{2pt}
\setlength{\belowdisplayskip}{1pt} 
\begin{split}
C_{\text{distance}}\left(\mathbf{x}_{1}, \mathbf{x}_{2}\right) & =\left|\mathbf{x}_{1} - \mathbf{x}_{2}\right|-d_0\\
C_{\text{volume}}\left(\mathbf{x_{1}, x_{2}, x_{3}, x_{4}}\right) & =\frac{1}{6}\left(\mathbf{x_{2,1}} \times \mathbf{x_{3,1}}\right) \cdot \mathbf{x_{4,1}} - v_0
\end{split}
\end{equation}

Where, $d_0$ is the initial distance between $\mathbf{x_1}\in \mathbb{R}^3$ and $\mathbf{x_2}\in \mathbb{R}^3$ and $v_0$ is the initial tetrahedron volume represented by the four corner particles $\mathbf{x_1,x_2,x_3,x_4}\in \mathbb{R}^3$.
The position corrections $\left[\Delta \mathbf{x}_i \right]_{\text{distance}}$ and $\left[\Delta \mathbf{x}_i \right]_{\text{volume}}$ can be obtained respectively.

\subsubsection{\textbf{Shape Matching}}
Shape matching is a geometrically motivated approach of simulating deformable objects \cite{Mller2005MeshlessDB} to preserve rigidity. The basic idea is to separate the particles into several local cluster regions and then, to find the best transformation that matches the set of particle positions (within the same cluster) before and after deformation, denoted by $\lbrace \mathbf{\hat{x}}_i \rbrace$ and $\lbrace \mathbf{x}_i \rbrace$, respectively.

The corresponding rotation matrix $\mathbf{R}$ and the translational vector $\hat{\mathbf{t}}$, $\mathbf{t}$ of each cluster are determined by minimizing the total error,
\begin{equation}
\setlength{\abovedisplayskip}{2pt}
\setlength{\belowdisplayskip}{2pt} 
{\arg \min}_{\mathbf{R}^*, \hat{\mathbf{t}}^*, \mathbf{t}^* } = \sum_{i}^{n} \| \mathbf{R}\left(\mathbf{\hat{x}}_{i}-\hat{\mathbf{t}}\right) + \mathbf{t} -\mathbf{x}_{i} \|_2^{2}
\end{equation}
where $n$ represents the number of particles in the corresponding cluster. The detailed solutions can be found in \cite{Macklin_2017} by polar decomposition of the transformation matrix. Thus, the position corrections of shape matching can be computed as
\begin{equation}
\setlength{\abovedisplayskip}{2pt}
\setlength{\belowdisplayskip}{2pt} 
\left[\Delta \mathbf{x}_i \right]_{\text{shape matching}} = \mathbf{R}^* \left(\mathbf{\hat{x}}_{i}-\hat{\mathbf{t}}^* \right) + \mathbf{t}^* -\mathbf{x}_{i}
\end{equation}

The above constraints can be computed through the Gauss-Seidel method \cite{Macklin_2017} (Lines 3 to 6 in Algorithm \ref{alg:PBD_simu}).
% \begin{equation}
% \Delta \mathbf{x}_i = \alpha \left[ \mathbf{R}^* \left(\mathbf{\hat{x}}_{i}-\hat{\mathbf{t}}^* \right) + \mathbf{t}^* -\mathbf{x}_{i} \right]
% \end{equation}
% where $\alpha$ is a user-defined stiffness parameter specify the rigidity.

% The main contribution of our work is providing a registration approach to optimize the differences between aforementioned PBD simulation and the real visual perception data. 
% In practice, simulation with dynamic solver can rarely represent the true deformation of an object, especially for soft tissues. 
% To evaluate the difference between simulation and camera observation, a cost function which penalize incorrect simulation is fundamental. Since the process of our algorithm is similar to traditional pbetween the simulationoint registration which aims to identify the same point in different frames, we name this algorithm as “Real-to-Sim Registration”.

\section{Registration}
\begin{figure*}[t!]
\vspace{2mm}
\centering
\includegraphics[width=0.99\textwidth]{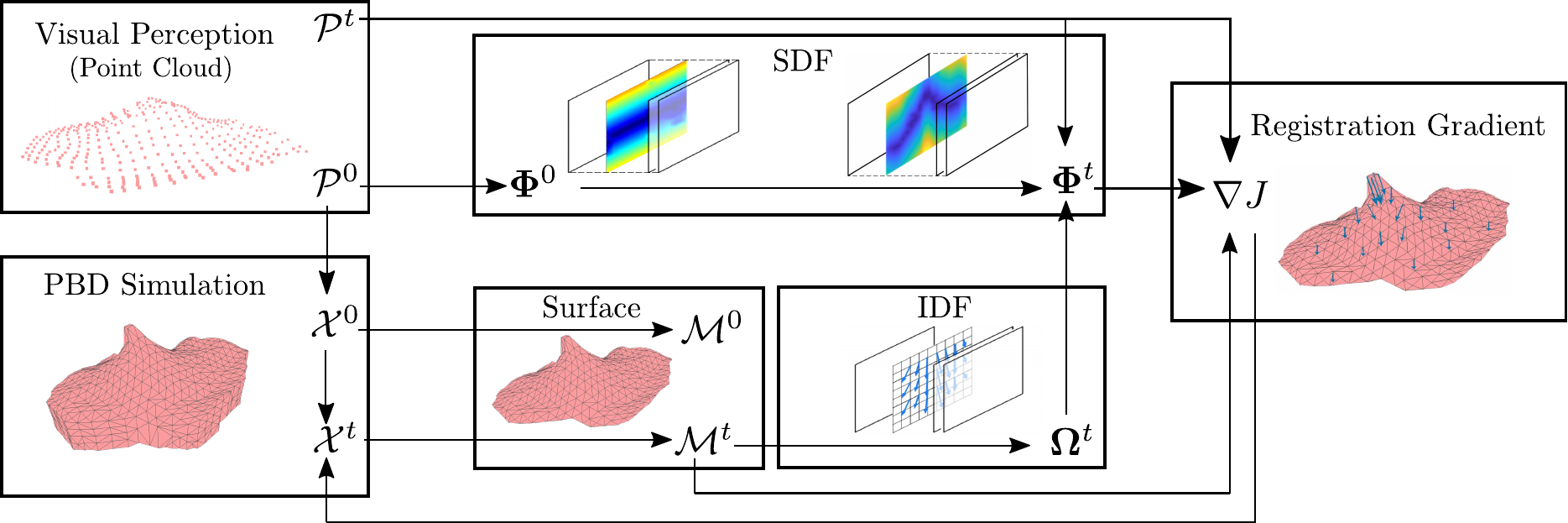}
\caption{The real-to-sim registration algorithm flow involves both the observed point cloud and PBD simulation. $\mathcal{P}^0$ and $\mathcal{P}^t$ are the observed point clouds at time $0$ and $t$ respectively. $\mathcal{X}^0$ and $\mathcal{X}^t$ are the simulated volume meshes (represented by particles) in PBD. 
$\mathcal{M}^0$ and $\mathcal{M}^t$ are the extracted surface (a subset of volume mesh) particles. $\mathbf{\Omega}^t$ is the inverse deformation field (IDF) of $\mathcal{M}^0$ described along a 3D grid. $\mathbf{\Phi}^0$ is the initial signed distance field (SDF) of $\mathcal{P}^0$ defined along the grid, and $\mathbf{\Phi}^t$ is the approximated SDF. The registration cost gradient $\nabla J^{t}$ can then be calculated for PBD simulation updates. The math symbols can also be referred to Algorithm \ref{alg:real_to_sim_registration}.
% Each subfigure demonstrated the concrete data visualization.
}
\label{fig:registration_flow}
\vspace{-0.2in}
\end{figure*}

To ensure our PBD simulator matches the real-world observations, we propose a real-to-sim registration algorithm, 
% It only requires a computation of the SDF field for visual observation at the initial frame.
% which requires to compute the signed distance function (SDF) field between the observed point cloud data and a deformed simulation. It leads to position correction as a registration cost.
which achieves position correction by minimizing a registration cost.
The main contribution of our work is 
bridging the gap between the PBD simulation and a 3D visual observation. In this work we will leverage the perception framework introduced in \cite{SuperLi2019}, which is a fusion-based method for surface reconstruction and deformable tissue tracking.
% \setlength{\textfloatsep}{2pt}
% \[
% \begin{minipage}{\displaywidth}
\begin{algorithm}[t!]
\SetAlgoLined
\SetKwInOut{Input}{input}
\SetKwInOut{Output}{output}
\Input{Initial point cloud $\mathcal{P}^{0}$, registration stiffness $\lambda_\text{regi}$} 
% \Output{Registration gradient $\nabla J^t$}
$\mathcal{X}^{0} \leftarrow \mathcal{P}^{0}$ \Comment{{\scriptsize initial PBD particles generation}} \\
$\mathbf{\Phi}^{0} \leftarrow \mathcal{P}^{0}$ \Comment{{\scriptsize initial SDF generation}} \\
$t = 0$\\
\While{not terminated}{
% $ \mathcal{X}^t \leftarrow \lbrace \mathbf{x}_1^t, \mathbf{x}_2^t, \cdots \rbrace$  \Comment{{\scriptsize perform PBD simulation}} \\
$ \mathcal{M}^t \leftarrow \mathcal{X}^t $  \Comment{{\scriptsize extract surface particles from PBD}} \\
$ \mathbf{\Omega}^{t} \leftarrow \mathcal{M}^t$  \Comment{{\scriptsize calculate inverse deformation field}} \\
$ \mathcal{P}^t \leftarrow \lbrace \mathbf{p}_1^t, \mathbf{p}_2^t, \cdots \rbrace$  \Comment{{\scriptsize get current point cloud}} \\
$ \mathbf{\Phi}^{t} \leftarrow \mathbf{\Phi}^{0}, \mathbf{\Omega}^{t}, \mathcal{P}^t $  \Comment{{\scriptsize approximate deformed SDF}} \\
$ J^t \leftarrow \mathbf{\Phi}^{t}, \mathcal{P}^t $\Comment{{\scriptsize calculate registration cost}} \\
$ \nabla J^t \leftarrow \mathcal{M}^t$ \Comment{{\scriptsize calculate registration gradient}} \\
$ \mathcal{X}^t \leftarrow \nabla J^t, \lambda_\text{regi}$  \Comment{{\scriptsize perform PBD simulation with registration}} \\
$t \leftarrow t + \Delta t$
}
\caption{Real-to-Sim Registration Flow}
\label{alg:real_to_sim_registration}
\end{algorithm}
% \lipsum[3-6]
% \setlength{\textfloatsep}{0pt}
% \end{minipage}
% \]

An outline of our real-to-sim registration is shown in Algorithm \ref{alg:real_to_sim_registration} and visualized in Fig. \ref{fig:registration_flow}. 
% A more comprehensive algorithm flow is shown in Fig. \ref{fig:registration_flow}. 
Signed distance function (SDF) field is used in this work to evaluate the difference between observed point cloud data and simulated deformation.
We firstly define the initial SDF field $\mathbf{\Phi}^{0}$, in a discrete Eulerian 3D space, using the first frame of reconstructed point cloud data $\mathcal{P}^{0}$ (detailed in Section \ref{sec:initial_SDF}). The SDF indicates the signed distance between a given space point %discrete spatial grid 
and the initial surface mesh constructed from point cloud $\mathcal{M}^{0}$.
%Meanwhile, the PBD simulation is initialized with the surface mesh with particle representation, according to this initial point cloud data. 
Meanwhile, the PBD simulation is also initialized using the initial surface mesh.
We extend the surface mesh into a volumetric tetrahedron mesh $\mathcal{X}^{0}$ along the gravity direction with pre-assumed thickness of the soft tissue.
% It plays as a reference configuration of the observation. 
Then at each time $t$, we construct an inverse deformation field (IDF)  $\mathbf{\Omega}^{t}$ by taking the simulated surface mesh $\mathcal{M}^{t}$  as input (detailed in Section \ref{sec:inverse_deformation_field}).
Thus, the deformed SDF $\mathbf{\Phi}^{t}$ of the point cloud $\mathcal{P}^{t}$ can be approximated by tracing back the surface  deformation using IDF $\mathbf{\Omega}^{t}$,
% Then we can approximate the deformed SDF $\mathbf{\Phi}^{t}$ of the tracked point cloud $\mathcal{P}^{t}$  by tracing back the surface mesh deformation $\mathbf{\Omega}^{t}$ as,
\begin{equation}
\setlength{\abovedisplayskip}{2pt}
\setlength{\belowdisplayskip}{2pt} 
    \mathbf{\Phi}^{t}(\mathcal{P}^{t}) \approx \mathbf{\Phi}^{0}(\mathcal{P}^{t} + \mathbf{\Omega}^{t}(\mathcal{M}^{t})) 
\label{eq:deformed_SDF}
\end{equation}

To sum up, only the SDF field at initial frame $\mathbf{\Phi}^{0}$ is computed and all other SDF values are estimated by tracing back with the IDF $\mathbf{\Omega}^{t}$. The SDF values are then evaluated as registration cost (detailed in Section \ref{sec_error}) and taken into PBD simulation.
\subsection{Initial SDF Generation}
\label{sec:initial_SDF}
\begin{figure}[t] 
    \centering
       \includegraphics[width=0.7\linewidth]{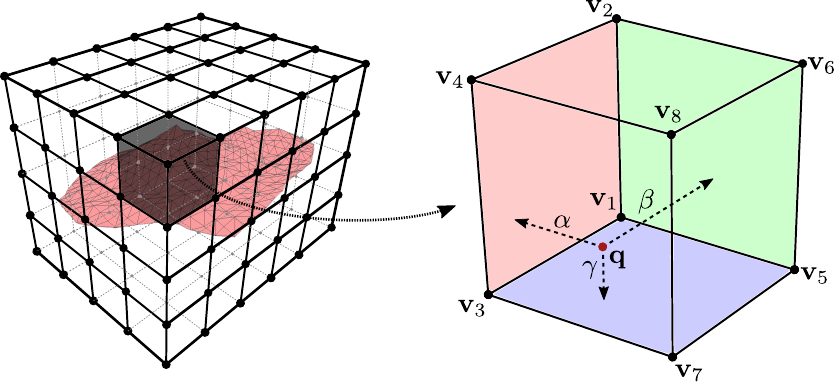}
  \caption{The boundary space (left) is discretized into Eulerian space grids. The simulated tissue surface is represented by mesh particles inside it. Each space point $\mathbf{p}$ is weighted by its 8 surrounding grid cube vertices $\mathbf{v}_i$  according to the normalized distance to each face.}
  \label{fig:eulerian_space_grid} 
  \vspace{-0.1in}
\end{figure}
%Drift may occur during the iterative approximation of the SDF using Eq. \ref{eq:deformed_SDF}. However, this is mainly due to the inverse deformation field calculation for whole Euler grid with diffusion of particles deformation, as detailed in the following Section \ref{sec:inverse_deformation_field}. We can assume the simulated particle deformation with incremental updates do not destroy the initial SDF property varied by time $t$.
% We can assume the local deformation difference between simulation and observation is relatively small. Meanwhile, the registration intervals is short.
In practice, instead of constructing a continuous 3D SDF field, we only discretize the boundary space that envelops the whole deforming mass into a 3D Eulerian grid $\mathbf{V}$, as shown in Fig.
\ref{fig:eulerian_space_grid}. Only the discrete SDF value at each grid vertex $\mathbf{v} \in \mathbb{R}^{3}$ is calculated. Then the SDF value of a position within (but not necessarily falling on) this grid can be identified via linear interpolation of its surrounding eight grid cube vertices ($\lbrace \mathbf{v}_{1}, \mathbf{v}_{2}, \mathbf{v}_{3}, \cdots, \mathbf{v}_{8} \rbrace$) in which the point resides. The interpolation weights are calculated according to the normalized distance to each face of the surrounding cube, named by $\lbrace \alpha, \beta, \gamma, 1-\alpha, 1-\beta, 1-\gamma \rbrace$. For any given point in space $\mathbf{q} \in \mathbb{R}^{3} $, the SDF vector is interpolated as 
\begin{equation}
\setlength{\abovedisplayskip}{3pt}
\setlength{\belowdisplayskip}{2pt} 
\begin{split}
\mathbf{\Phi}^{0}(\mathbf{q})  
& = (1-\alpha)(1-\beta)(1-\gamma)\mathbf{\Phi}^{0}(\mathbf{v}_1) \\
& + (1-\alpha)(1-\beta) \gamma \mathbf{\Phi}^{0}(\mathbf{v}_2) \\
& + (1-\alpha) \beta (1-\gamma) \mathbf{\Phi}^{0}(\mathbf{v}_3) + ... +\alpha \beta \gamma \mathbf{\Phi}^{0}(\mathbf{v}_8)\\
\end{split}
\label{eq:linear_interpolation_grid}
\end{equation}
where $\mathbf{\Phi}^{0}(\mathbf{v}_i), i \in \lbrace 1, 2, \cdots, 8 \rbrace$ is the initial SDF vector for its surrounding grid vertices. This is calculated by the distance to corresponding closest point $\mathbf{p}_*^{0} \in \mathbb{R}^{3}$ inside the received initial point cloud frame $\mathcal{P}^{0}$. Then, the initial SDF vector for each grid vertex $\mathbf{v} \in \mathbf{V}$ is defined by 
\begin{equation}
\setlength{\abovedisplayskip}{3pt}
\setlength{\belowdisplayskip}{3pt} 
\begin{split}
&\mathbf{\Phi}^{0}(\mathbf{v})  = \mathbf{v} - \mathbf{p}_*^{0} \\
& \mathbf{p}_*^0 = {\arg \min}_{\mathbf{p}^0 \in \mathcal{P}^{0}} \| \mathbf{v} - \mathbf{p}^{0}\|_2 
% & \mathbf{\Phi}^{0}(\mathbf{v}_i) = \| \mathbf{v}_i - \mathbf{p}_*^{0}\|^2
\end{split}    
\end{equation}

\subsection{Inverse Deformation Field (IDF)}
\label{sec:inverse_deformation_field}

\begin{figure}[t]
\vspace{2mm}
    \centering
       \includegraphics[width=0.99\linewidth]{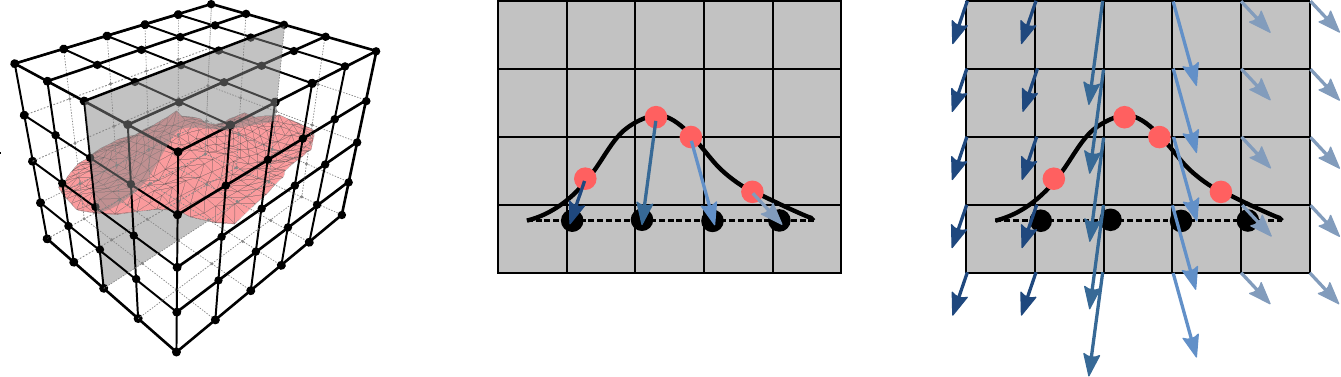}
  \caption{The demonstration of inverse deformation field (IDF). For simplicity, we visualize the 3D IDF (left) by one 2D Eulerian grid slice (middle and right). 
  Middle figure shows the inverse deformation vectors of current surface particles (red points on solid curve) regarding initial particles (black points on dashed curve). Right figure shows the diffusion of whole grid vertices.}
%   The first step (middle), for each current Lagrangian surface mesh particles (red points on solid curve), we calculate the inverse deformation vectors regarding the initial surface particles (black points on dashed curve). The second step (right) is diffusion of these inverse deformation vectors. For each grid vertex, it is assigned as the same inverse deformation vector t he one 
  \label{fig:inverse_deformation_field} 
  \vspace{-0.1in}
\end{figure}

%An inverse deformation field (IDF) can be computed by tracing back the simulated particle-wise displacements to their initial positions, as shown in Fig.
An inverse deformation field (IDF) can be computed by tracing back the positions of particles to their initial ones, as shown in Fig.
\ref{fig:inverse_deformation_field}. First, for each surface particle $\mathbf{m}^t_i \in \mathbb{R}^{3}$ in the surface mesh set $\mathcal{M}^{t}$ at current time $t$, we can obtain the deformation vector by subtracting the corresponding particle at time $t=0$, $\mathbf{m}^0_i \in \mathbb{R}^{3}$ as,
\begin{equation}
\setlength{\abovedisplayskip}{3pt}
\setlength{\belowdisplayskip}{3pt} 
\begin{split}
&\mathbf{\Omega}^{t}(\mathbf{m}^t_i)  = \mathbf{m}^0_i - \mathbf{m}^t_i \\
% & \mathbf{\Phi}^{0}(\mathbf{v}_i) = \| \mathbf{v}_i - \mathbf{p}_*^{0}\|^2
\end{split}    
\end{equation}
where $\mathbf{m}^t_i$, $\mathbf{m}^0_i$ can be acquired directly from the PBD simulation.
% at each time $t$.

%Then, for all of the Eulerian grid vertices $\mathbf{v} \in \mathbf{V}$ discretized in the space as the same one during initial SDF generation, we can define their corresponding deformation field vector as,
Then, for all of the discreate Eulerian grid vertices $\mathbf{v} \in \mathbf{V}$ in the initial SDF space, we define their corresponding deformation field vector as,
\begin{equation}
\setlength{\abovedisplayskip}{3pt}
\setlength{\belowdisplayskip}{3pt} 
\begin{split}
&\mathbf{\Omega}^{t}(\mathbf{v})   = \mathbf{\Omega}^{t}(\mathbf{m}^t_*) \\
& * = {\arg \min}_{\mathbf{m}_*^0 \in \mathcal{M}^{0}} \| \mathbf{v} - \mathbf{m}_*^{0}\|_2 \\
% & \mathbf{\Phi}^{0}(\mathbf{v}_i) = \| \mathbf{v}_i - \mathbf{p}_*^{0}\|^2
\end{split}
\label{eq:inverse_deformation_field}
\end{equation}
%where $*$ is the index of the closest initial surface particle in initial surface set $\mathcal{M}^{0}$, with respect to the grid vertex $\mathbf{v}$. 
where $*$ is the index of the closest particle to the grid vertex $\mathbf{v}$ in initial surface set $\mathcal{M}^{0}$. 
It can be viewed as a diffusion operation for each Eulerian grid vertex. 
%Research on level set methods have shown that such extensions are only acceptable in a narrow band around the set of surface particles \cite{osher2004level}, but this is reasonable in situation since our simulated deformations are relative small for surgical scenarios (in particular for soft tissue with scale of millimeter), and the time interval between two frames of registration is relatively short. Therefore, we can assume our simulated particle deformation lies in the narrow band around the observed point cloud data.
%Then, the final step is to extend the deformation field into the whole space for any given point $\mathbf{q} \in \mathbb{R}^{3}$. This is done by using the same interpolation method involving a weighted combination of the deformation field at the surrounding eight grid vertices given in Eq. \ref{eq:linear_interpolation_grid}.
For any other point  $\mathbf{q} \in \mathbb{R}^{3}$ in the initial SDF space, the deformation field is calculated using a similar interpolation method as the one shown in  Eq. \ref{eq:linear_interpolation_grid}.

% The last problem is to calculate the displacement at each grid vertex. By comparing the observed point cloud data, we can easily compute the displacement at surface particles. Hence, we need to extend the displacement of surface into the whole space, which is actually a diffusion problem. This problem is solved in Eulerian space by matching the grid vertex with its closest surface particle. 

% (figure 2: show the correspondence between frames) 

\subsection{Real-to-Sim Registration Cost}
\label{sec_error}
In this section, we define the real-to-sim registration cost function. This registration refers to the matching between the immediate visual perception and the PBD simulation of the current timestamp. The matching cost can be defined as the summation of the deformed SDF values approximated using the surface mesh particles $\mathcal{M}^{t}$ and all visual perception data, i.e. point cloud $\mathcal{P}^{t}$. Suppose $\mathbf{p}_i^{t} \in \mathbb{R}^{3}$ is the $i$-th point in $\mathcal{P}^{t}$, then the registration cost function is formulated as
% Suppose $\mathbf{x}_i^{t} \in \mathbb{R}^{3}$ is each surface particle simulated in PBD inside $\mathcal{M}^{t}$, our registration cost function can be formulated as
\begin{equation}   
\setlength{\abovedisplayskip}{2pt}
\setlength{\belowdisplayskip}{2pt} 
J^t(\mathcal{M}^t) = \sum_{i=0}^{n} \| \mathbf{\Phi}^{t}(\mathbf{p}_i^{t}) \|^2_2, ~ \mathbf{p}_i^{t} \in \mathcal{P}^{t}
\label{eq:registration_cost}
\end{equation}
where $n$ is the number of points in the reconstructed point cloud $\mathcal{P}^{t}$.

Since our algorithm consists of multiple discrete grid calculations which precludes analytical gradients, the registration process is performed as a numerical gradient descent via a backwards difference approach, %The partial derivatives of registration cost with respect to each surface particle $\mathbf{m}^t \in \mathcal{M}^t$ at time $t$ are calculated numerically. 
%This is because our algorithm consists of multiple discrete Euler grid calculation.% it's difficult to generate an analytical function of derivatives. 
%Hence, a finite difference method is used, 
\begin{equation}   
% \frac{\partial J^t}{\partial \mathbf{m}^t} = \frac{J^t(\mathbf{m}^t + \Delta \mathbf{m}) - J^t(\mathbf{m}^t) }{\Delta \mathbf{m}}
\nabla_{\mathbf{m}^t} J^t = \frac{J^t(\mathbf{m}^t + \Delta \mathbf{m}) - J^t(\mathbf{m}^t) }{\Delta \mathbf{m}}
\label{eq:registration_gradients}
\end{equation}
where $\Delta \mathbf{m} \in \mathbb{R}^3$ is a manually assigned small forward deformation of surface particles.
%gradient with a small $\Delta x$. In this way, PBD solver can find the direction to reduce registration error. 

\subsection{Constraint Satisfaction for Real-to-Sim Registration} 

% Since PBD solver is a gradient-driven module, the derivatives of registration error are needed.
% Since PBD simulation use G
Traditional point-to-point registration will force all particles on the surface to the observed position, which may violate the object's geometrical structure if the tracking algorithm provides an incorrect correspondence. To avoid this, we perform correspondence-free corrections by minimizing the difference between two surfaces instead of pairs of corresponding points. 
Since point-to-point correspondences are not strictly enforced, the error can hardly be zero.
% As a result, the error can never be zero and point-to-point correspondences are not strictly enforced.
However, by pulling each simulated particle along the total registration gradient, the points will finally rest in a neighborhood of the observation and consist of a similar tissue surface.

%Our real-to-sim registration can be seen as a visual perception-based surface constraint, which can be directly integrated in the PBD simulation. All the simulated surface particles $\mathcal{M}^t$ are associated with this constraint. 

In Eq. \ref{eq:registration_cost} and \ref{eq:registration_gradients}, the summation of registration cost $J^t$ and the gradients of each surface particle $\nabla_{\mathbf{m}} J^t$ are obtained, which correspond to another constraint $C$ and $\nabla_{\mathbf{x}} C$ in Algorithm \ref{alg:PBD_simu}, respectively. Thus, the position correction introduced by real-to-sim registration can be directly updated as an additional, soft constraint. We introduce a stiffness parameter $\lambda_\text{regi} \in [0,1]$ to tune this constraint:
\begin{equation}
\setlength{\abovedisplayskip}{2pt}
\setlength{\belowdisplayskip}{2pt} 
\left[\Delta \mathbf{x} \right]_{\text{registration}} = \lambda_\text{regi} \cdot \nabla_{\mathbf{m}} J^t, ~~ \mathbf{x} = \mathbf{m}
\end{equation}
With the stiffness term $\lambda_\text{regi}$, the simulator will not force the surface immediately to the observed point cloud, which avoids oscillation while trying to satisfy different constraints in Gauss-Seidel style.

\section{Experiments and Results}
% \begin{figure}[t!] 
% \setlength{\belowcaptionskip}{-0.4cm} 
%     \centering
%     \includegraphics[width=0.3\linewidth]{mask/real.jpg}
%       \includegraphics[width=0.3\linewidth]{mask/pointcloud.jpg}
%       \includegraphics[width=0.3\linewidth]{mask/simulation.jpg}       
%   \caption{An example of the inaccurate reconstructions of the regions occluded by the surgical tool. The left figure is the real scene. The middle figure is the observed point cloud and the right figure is the simulation result without the registration. The red circle denotes the occluded regions in each figure. It is obvious that the estimation of the occluded region is inaccurate, which makes it reasonable to exclude these occluded points from our overall error measurements.}
%   \label{fig:exp_cam} 
% \end{figure}

\begin{figure}[t!]
\vspace{2mm}
\begin{subfigure}{0.24\textwidth}
\includegraphics[width=1\textwidth,height=1.0in]{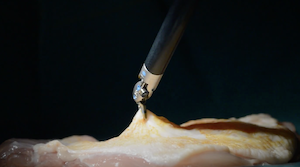}
\vspace{-0.12in}
\end{subfigure}
\begin{subfigure}{0.24\textwidth}
\includegraphics[width=1\textwidth,height=1.0in]{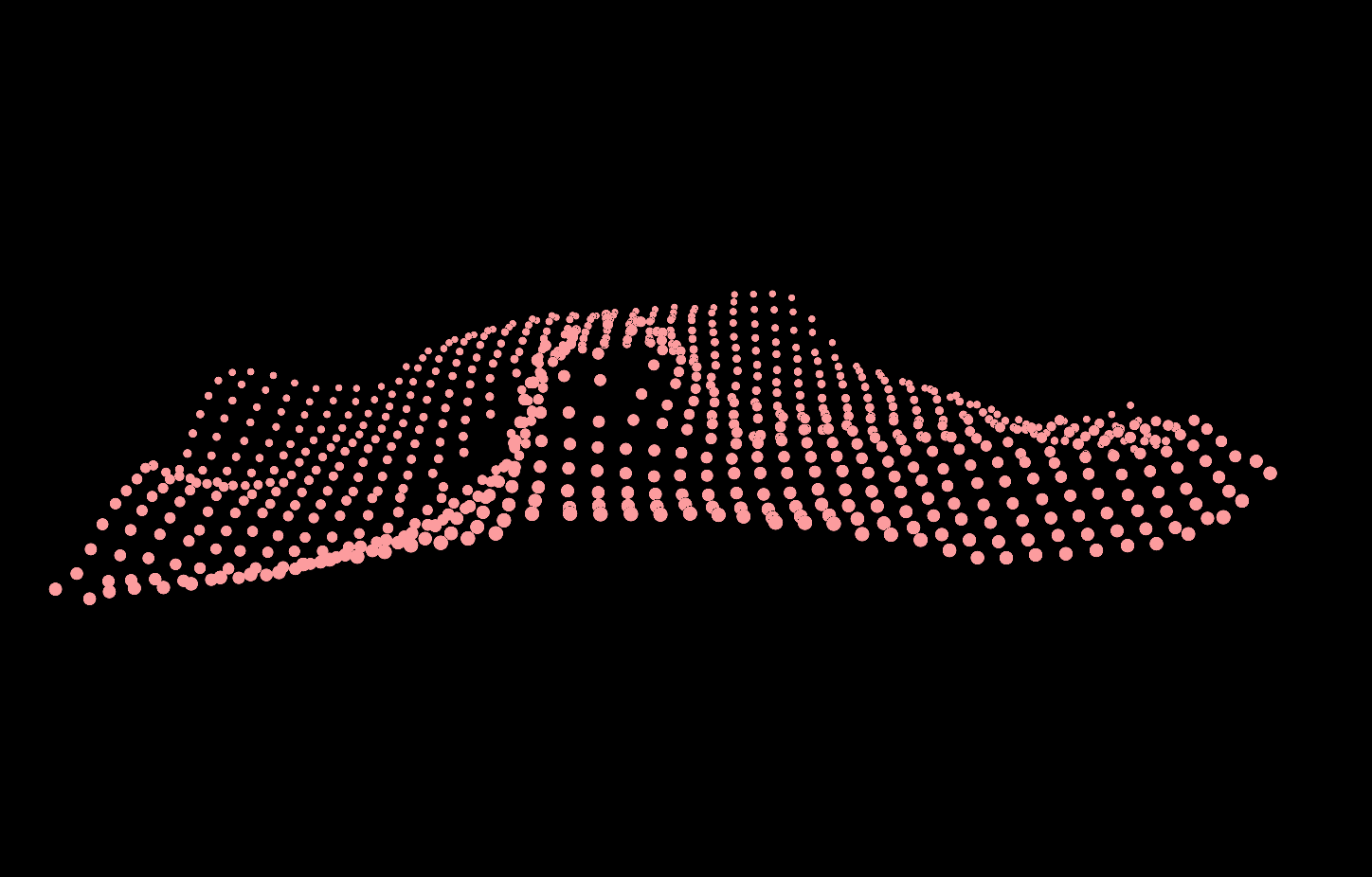}
\vspace{-0.12in}
\end{subfigure}
\begin{subfigure}{0.24\textwidth}
\includegraphics[width=1\textwidth]{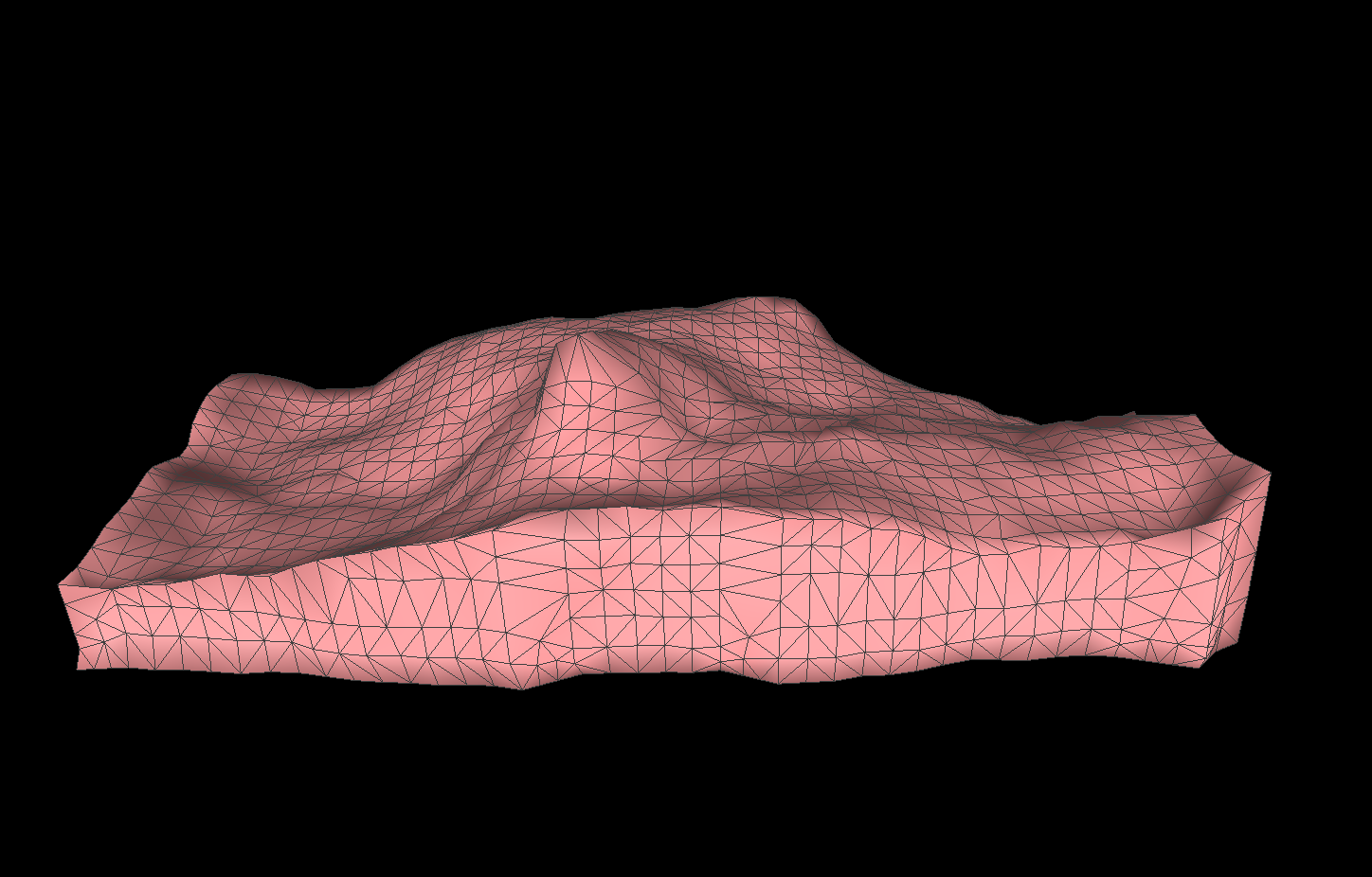}
\end{subfigure}
\begin{subfigure}{0.24\textwidth}
\includegraphics[width=1\textwidth]{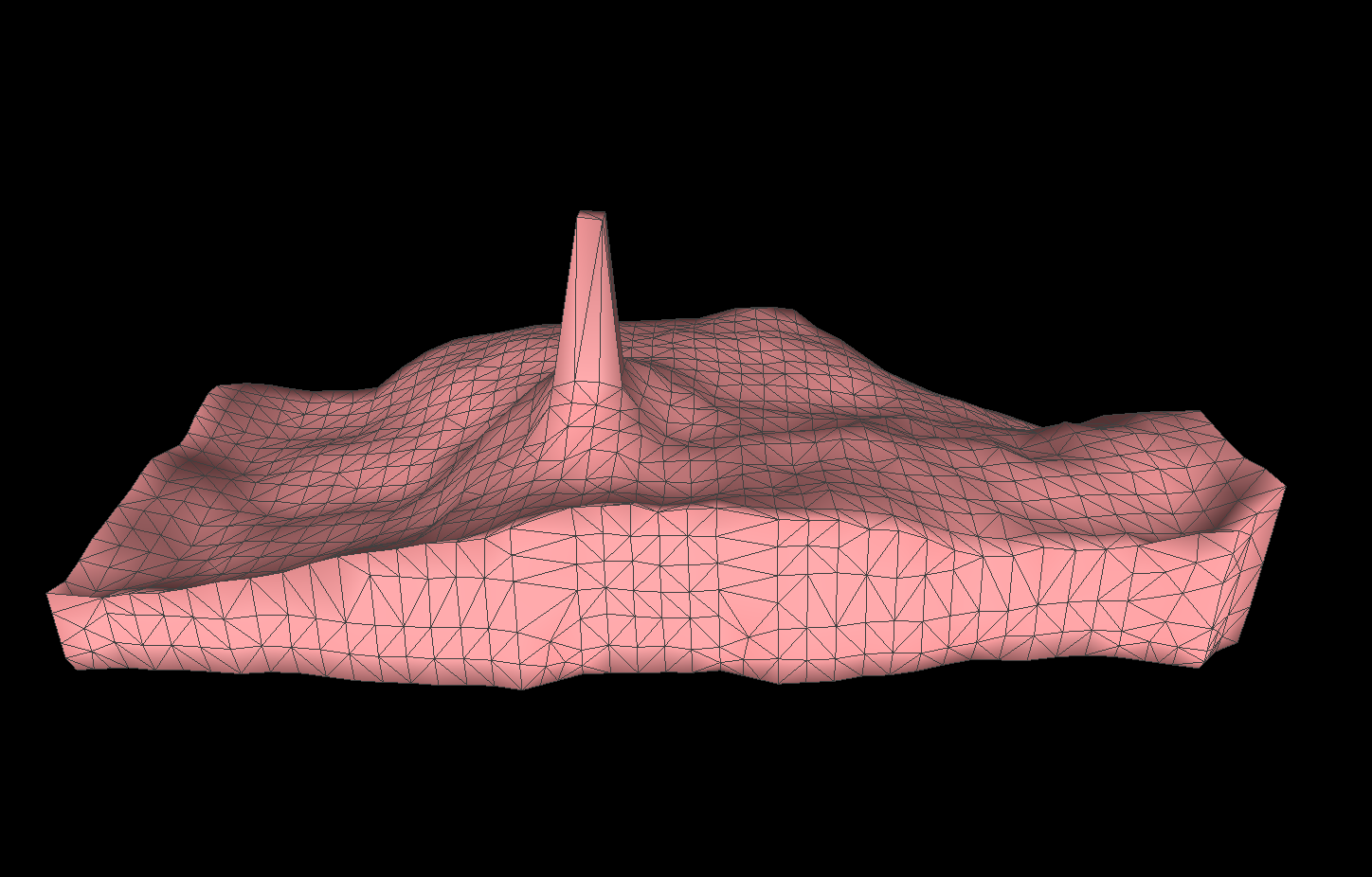}
\end{subfigure}
\caption{The demonstration of reconstructed volume meshes on the tissue manipulation dataset. The top-left figure shows the real image and the top-right figure shows the tracked surface particles using the surgical perception framework. The bottom figures show the simulation results, with (left) and without (right) the registration. The estimated mesh is more realistic after the real-to-sim registration.}
\label{fig:chicken_exp}
\vspace{-0.1in}
\end{figure}

\begin{figure}[t!] 
    \centering
       \includegraphics[width=0.48\linewidth, trim={0 0 11mm 0},clip]{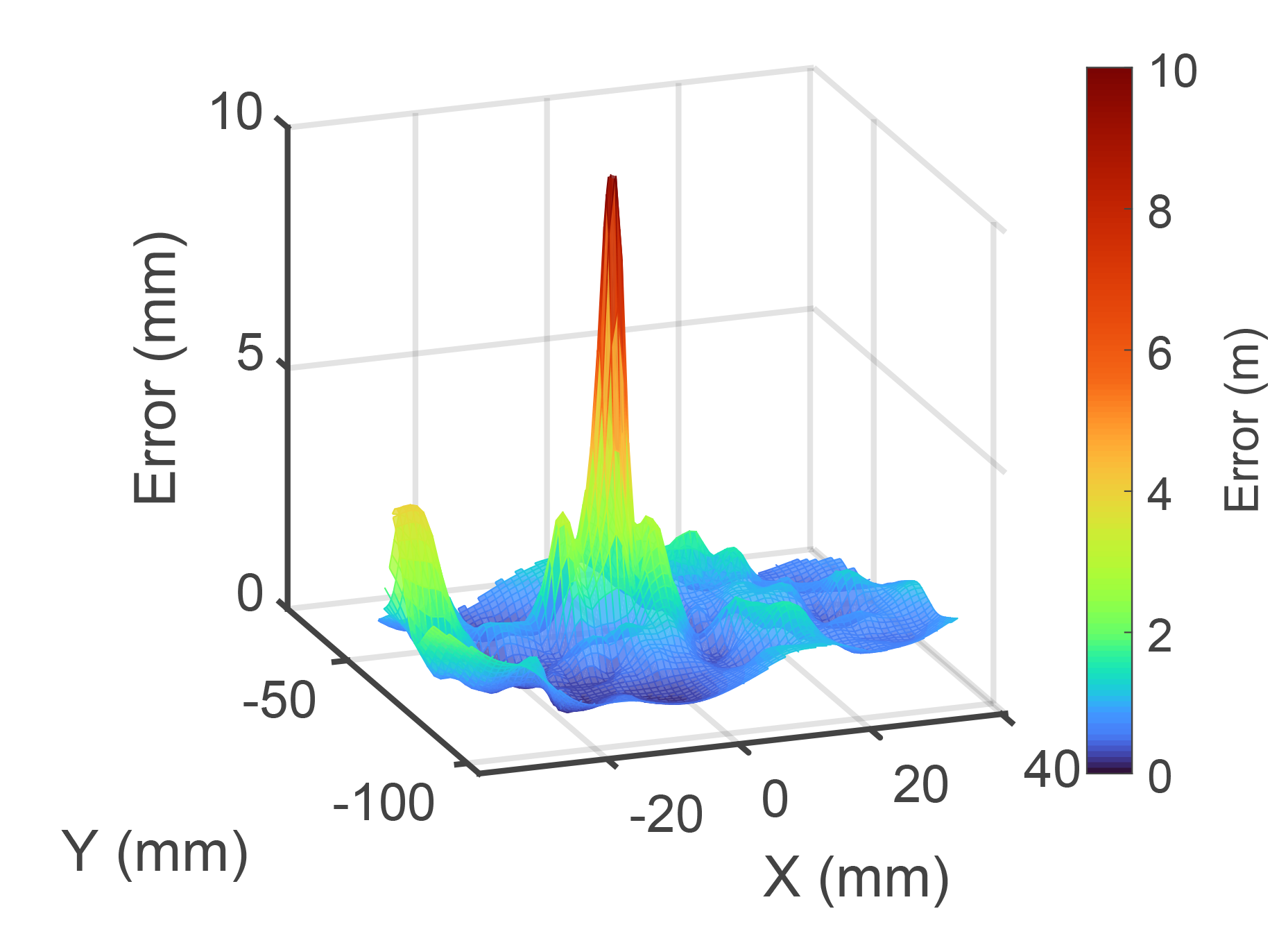}
        \includegraphics[width=0.48\linewidth, trim={0 0 11mm 0},clip]{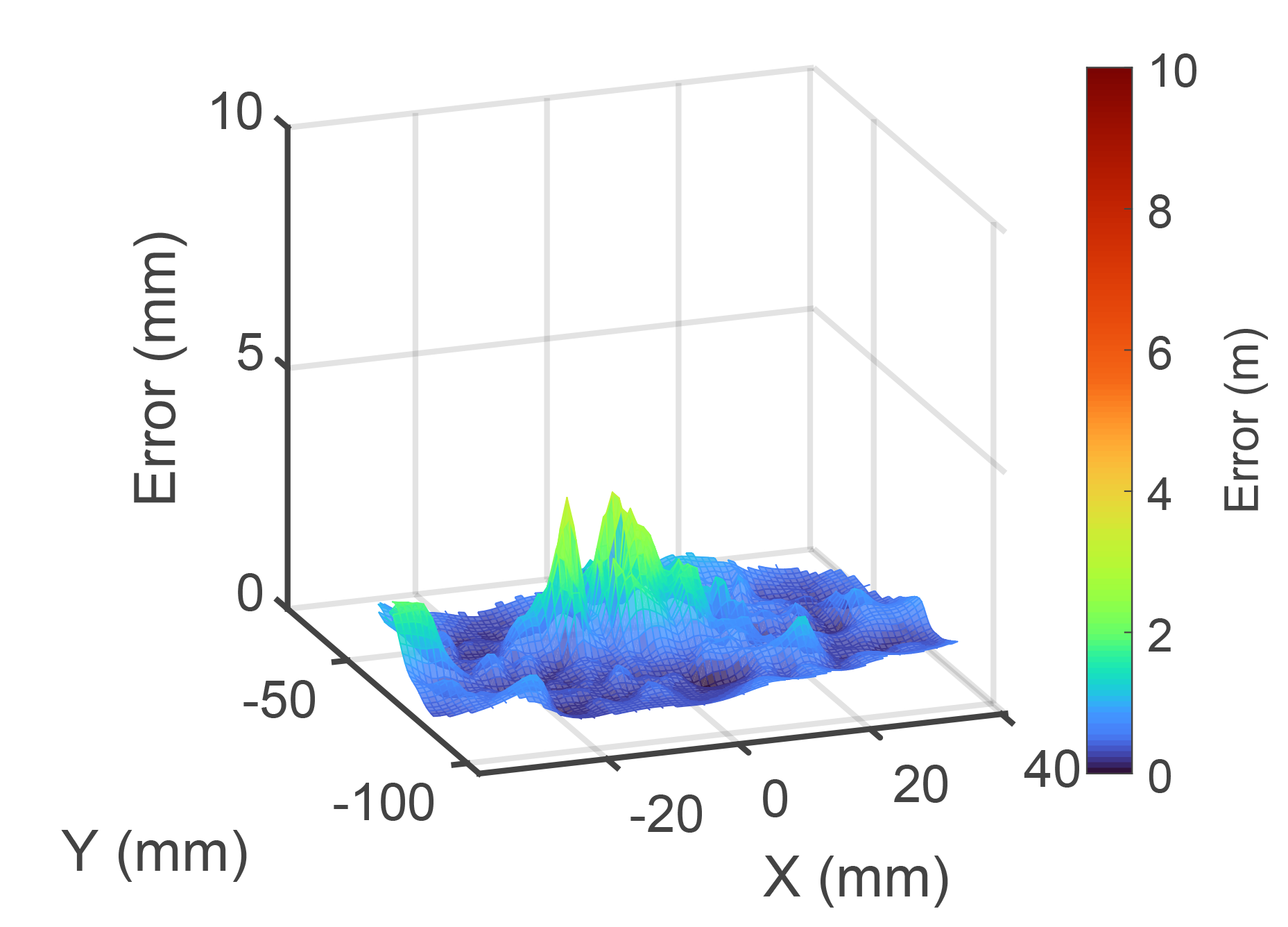}
  \caption{The visualization of the error between the simulation and observed surface particles before (left) the after (right) the real-to-sim registration (averaged in timestamp). After the registration, the real-to-sim error is reduced significantly around the pinch point.}
  \label{fig:chicken_error_mesh} 
  \vspace{-0.1in}
\end{figure}

%LineCharts -> get rid of xy plane
\begin{figure}[t] 
    \centering
       \includegraphics[width=0.9\linewidth]{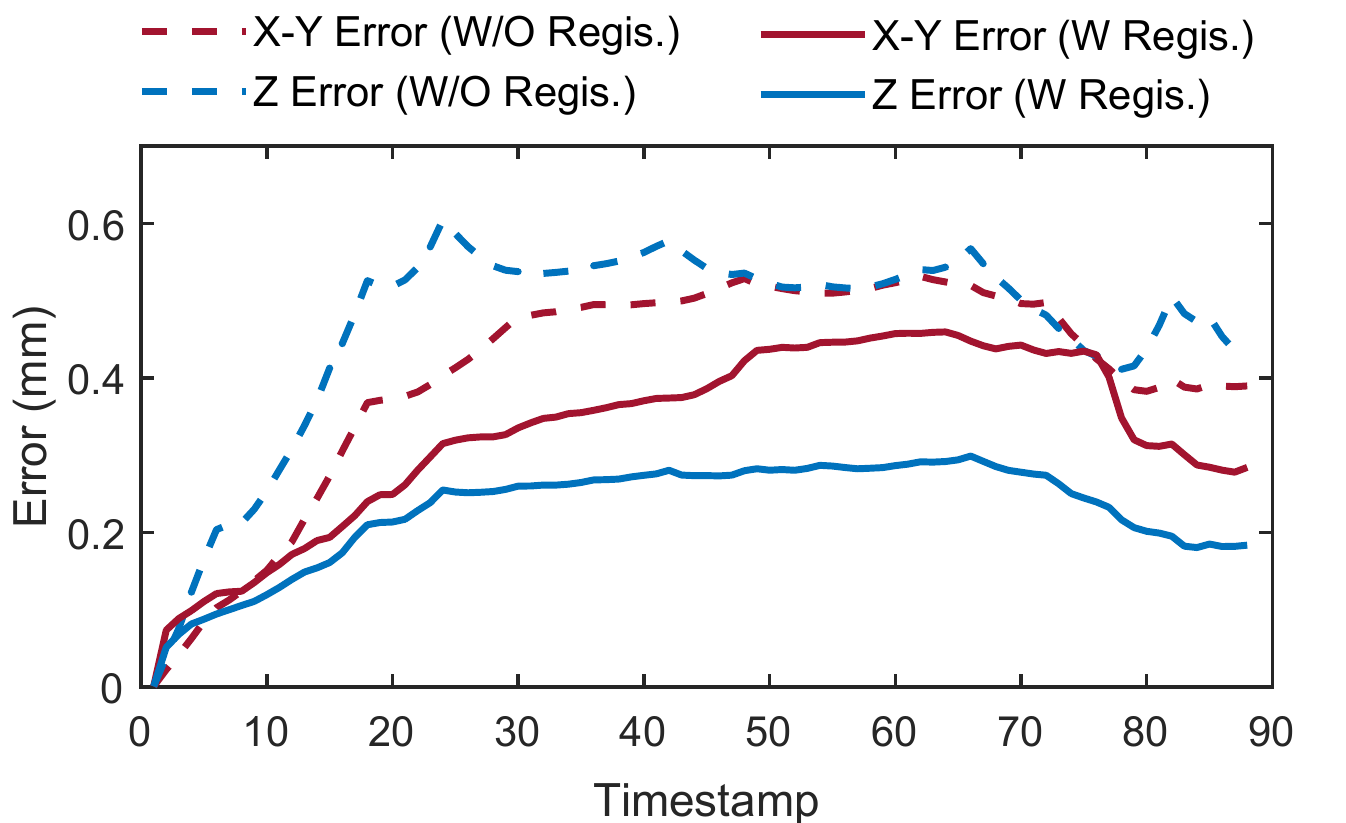}
  \caption{The real-to-sim registration errors (averaged over all surface particles) on $XY$-plane and on $Z$-direction (gravity direction). The errors both in the $Z$-direction and $XY$-plane are decreased after registration.
  }
  \label{fig:chicken_error_line} 
  \vspace{-0.1in}
\end{figure}
%   since we do not consider the tissue's hysteresis (can observed ) caused by the stretched deformation in $XY$-plane shown in the dashed block. 

\subsection{Experiment Setups and Evaluation Metrics}

In order to demonstrate the effectiveness of the proposed registration framework, we conducted experiments on two different live environments involving soft tissues manipulated using the da Vinci Research Kit (dVRK \cite{Kazanzides_2014_dvrk}): (1) the Chicken Skin Experiment from SuPer dataset \cite{SuperLi2019}, and (2) the Pork Steak Experiment, which consists of four motion trajectories: lift, cube, butterfly and sine wave.
For each experiment, the visual perception framework \cite{SuperLi2019} was utilized to track the tissue surface point cloud as the real-world observation after masking out the background area. The volume meshes (represented by particles) were created from the initially reconstructed point cloud before the manipulation, and the PBD simulation process started as the control actions were executed. 

The control actions involved grasping the surface of the tissue and producing a tissue deformation to track for the real-to-sim method. The grasp location was defined in the simulation by the four closest surface particles to the end-effector.
% pinch point (where the tool-tissue interaction happens). %Their positions were only determined by the gripper's position without corrections from the geometrical constraints.
During the registration, their positions were corrected using the shape matching method with the observed point cloud. The simulation boundary conditions were satisfied by fixing the boundary particles' position from the initial volume mesh, where the real chicken skin and pork steak were fixed on the table. This would be representative of an internal cavity where tissue would not typically be separated from connected organ before cutting.
% due to complications that might arise from adhesions \cite{van2018adhesion}. 

% % Vote for not mentioning this one ######### 
% The registration stiffness parameter $\lambda$ of real-to-sim registration in each experiment is described in Table  \ref{tab:simu_para}.
% \begin{table}[H]
% \centering
% \begin{tabular}{ccccc}
% \toprule  
% \text{Chicken skin} & \text{Lift} & \text{Cube} & \text{Butterfly} & \text{Sine Wave}\\
% \midrule  
% 0.5 & 0.1 & 0.1 & 0.2 & 0.15\\
% \bottomrule 
% \end{tabular}
% \caption{Registration stiffness parameter $\lambda$ for different experiments}
% \label{tab:simu_para}
% \end{table}
% %Error definition  heatmap, linear, box
\begin{figure*}[t]
\vspace{2mm}
\setlength{\belowcaptionskip}{-0.4cm} 
\centering
\includegraphics[width=1\linewidth]{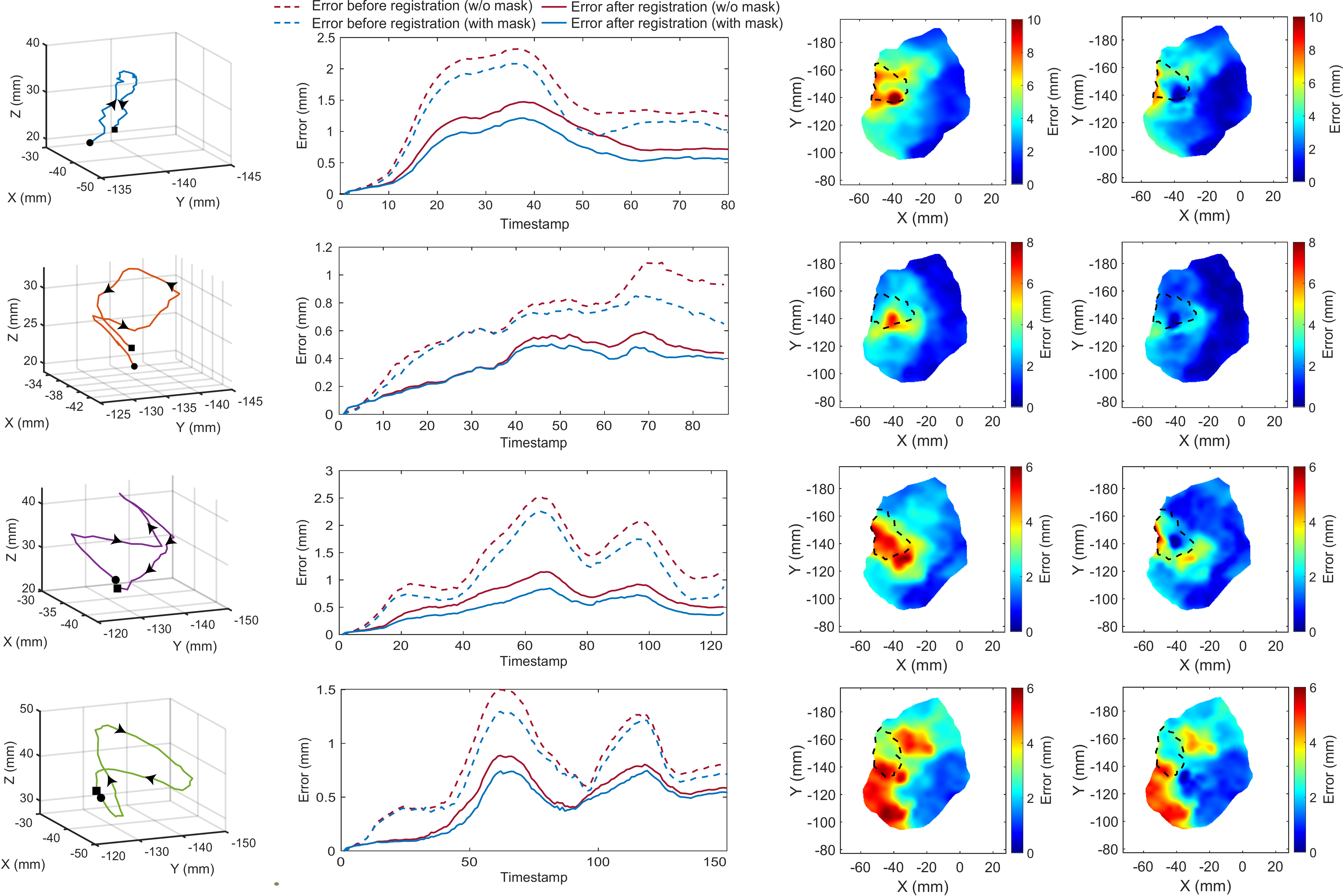}
  \caption{The quantitative results of the proposed real-to-sim registration method for four different manipulations (one for each row) in Pork Steak Experiment. The plots in the first column show the real tool trajectories (lift, cube, butterfly, and sine wave from top to bottom, respectively). 
  The second column shows the plots of real-to-sim errors before and after registration in time by averaging the surface particles (with and without masking of the occluded particles). The third and fourth columns show the real-to-sim errors in space (averaged over the timestamps) with and without registration respectively. The areas circled by dashed lines indicate the regions occluded by the tool. Our method significantly reduced errors in different manipulation tasks.}
  \label{fig:traj} 
\end{figure*}

\begin{figure}[t!] 
\setlength{\belowcaptionskip}{-0.4cm}
    \centering
    \includegraphics[width=0.3\linewidth]{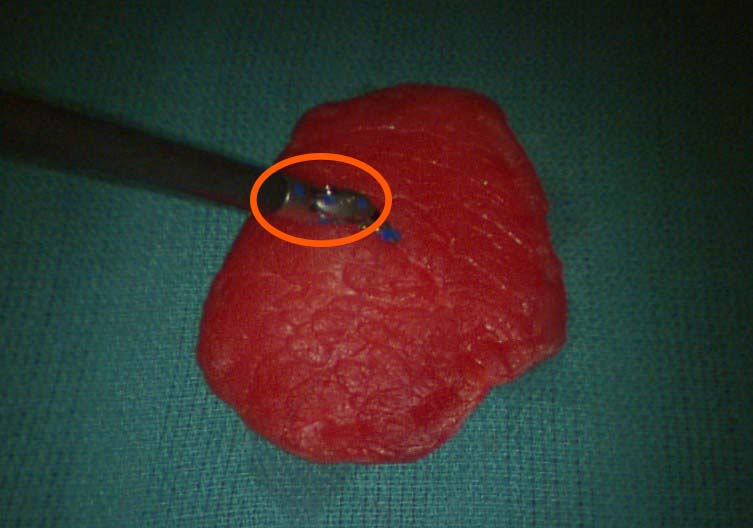}
       \includegraphics[width=0.3\linewidth]{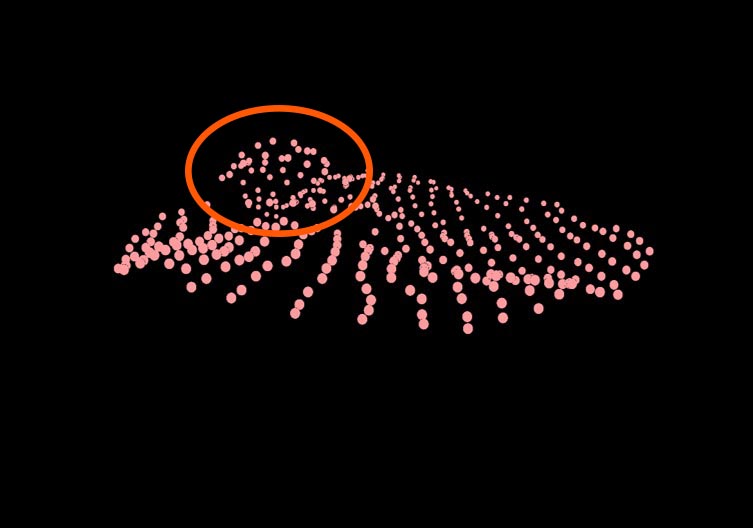}
       \includegraphics[width=0.3\linewidth]{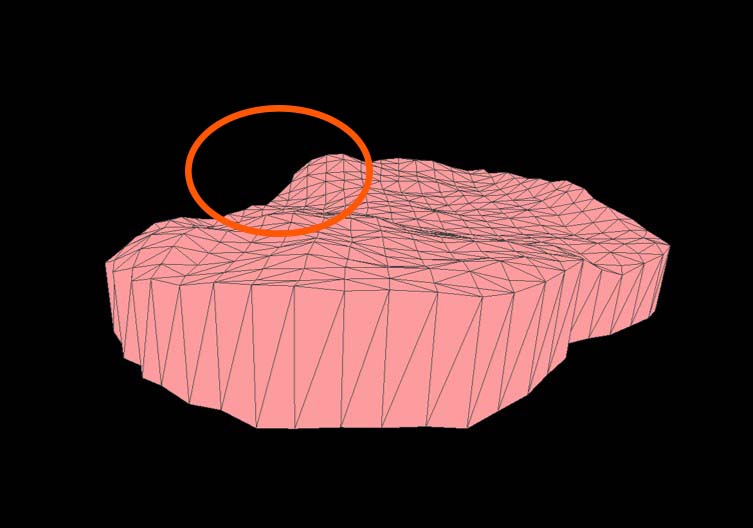}       
  \caption{An example of the inaccurate reconstructions of the regions occluded by the surgical tool. The left figure is the real scene. The middle figure is the observed point cloud and the right figure is the simulation result without registration. The orange circles indicate the occluded regions in each figure. It is obvious that the observation of the occluded regions is inaccurate.}
  \label{fig:exp_cam} 
  \vspace{-0.1in}
\end{figure}

In order to quantitatively evaluate the system performance, we evaluated the registration cost of both whole surface and individual surface particles. 
% i.e., $[\| \mathbf{\Phi}^{t}(\mathbf{p}_i^{t}) \|_2]$, ~ \mathbf{p}_i^{t} \in \mathcal{P}^{t}]$
We define the evaluation metric as the $L_2$ norm between the 3D position of the observed point cloud $\mathbf{p}_i^{obs} \in \mathcal{P}^t$ and the corresponding simulated surface particle $\mathbf{m}_i^{sim} \in \mathcal{M}^t$:
\begin{equation}\label{err}
\setlength{\abovedisplayskip}{1pt}
\setlength{\belowdisplayskip}{1pt} 
    \text{Error}_{\text{with/without regis}} = \sum_{i}^{n}\left\|\mathbf{m}^{sim}_i - \mathbf{p}_{i}^{obs}\right\|_2
\end{equation}
where, $n$ is the total number of simulated surface particles. It is necessary to mention that the evaluation metric defined here is different from the registration cost (which is using SDF) in the previous section, and will be averaged over timestamps or number of particles in the following data analysis. The surface particles are from PBD simulation at each time $t$. Both the simulation cost with registration and the one without are documented.
% which is optimized using the registration and satisfying other geometrical constraints.
% metric is defined as the sum of the simple $L_2$ norm for a better comparison of the results before and after the real-to-sim registration.

\subsection{Chicken Skin Experiment}
In this experiment, a surgical tool with a gripper was used to lift the chicken skin, as shown in Fig. \ref{fig:chicken_exp}. If we performed PBD without our registration method, the simulated volume mesh would not deform to the same shape as visually observed. The quantitative comparison results shown in Fig. \ref{fig:chicken_error_mesh} also support our observation.
After performing real-to-sim registration, PBD simulation was able to capture the surface deformation as observed from point cloud.
% our method is able to capture the surface deformation without destroying the topology of the simulated tissue.
In the left of Fig. \ref{fig:chicken_error_mesh}, the errors around grasping areas is abnormally high due to lack of realistic tissue parameters in simulation, while in the right figure, our method significantly reduce the error between simulation and observation.
% which may result from inaccurate tool trajectory estimation.
From Fig. \ref{fig:chicken_error_line}, we can tell that our method corrects both the errors in the $Z$-direction (gravity direction) and in $XY$-plane.
However, the $XY$ errors remain large even after registration.
It is caused by the uncertainties, i.e., noises from stereo reconstruction, tracking noises of the surgical tool etc.
Meanwhile, the deformation is mostly happening in anti-gravity direction ($Z$) in our grasping experiments, while only small deformation (in millimeter level) is presented in $XY$-plane.
% The impact of noises is more visible compared to $XY$ deformation.
The noise is relatively large comparing to $XY$ deformation and undermines the real result. Hence, we will focus on the error introduced in the $Z$-direction in the following experiments. 

% hysteresis of tissue,
% since we do not consider the tissue's hysteresis caused by the deformation.
% Thus, it doesn't 
% Meanwhile, the deformation is mostly happening in anti-gravity direction ($Z$) considering our grasping experiments.

%It is of interest to point out that in this experiment, the lifting were executed along the anti-gravity direction ($z$ axis), which means the dominant error component should be along $z$ axis. However, the results show that the error components on $XY$ plane are equal to or even larger than it along $z$ axis. Besides, after the tool went back to the start point, the error components on $XY$ plane still exists. The possible reason is the hysteresis of the deformation process. Hence, in the next experiment sets with more complex tool's trajectory, the error are projected along $z$ axis.

% \def \trimFactor {0.22}

\subsection{Pork Steak Experiment}
In this experiment, we tested our method by manipulating the tissue with four different moving trajectories, which are shown in the first column of Fig. \ref{fig:traj}. The following columns show the plots of real-to-sim errors in time (averaged over all surface particles) and the heatmaps of the real-to-sim errors in space (averaged over the timestamps) with and without registration, repectively. The experiment results show the importance of online, real-to-sim registration in properly representing the scene deformation.
%%fusion mask

The areas circled by the black dash lines in the heatmaps are the regions occluded by the surgical tool during the manipulation (see Fig. \ref{fig:exp_cam}). This information is typically not available, but we were using the SuPer framework \cite{SuperLi2019} for reconstruction, which does spatio-temporal fusion under partial occlusions to estimate their position. Because these points are being only estimated and not measured in the video frame, we can exclude those occluded points from our overall error measurements. %Therefore, the real-to-sim error of this region doesn't reflect the true scenario. To exclude the particle in that region, we reprojected all the observed point cloud into 2D images and check whether they fall into the surgical tool areas.
The second column in Fig. \ref{fig:traj} shows the mean real-to-sim registration errors (with and without mask) in time. The red solid line shows the error with registration averaging over all surface particles, while the blue solid line shows the averaged error with registration after excluding the particles that are occluded for more than half of the total frames. %We can see that the average registration errors are decreased by masking out those occluded particles in all scenarios. 
Since the SuPer framework deals with the occluded area using the history information for fusion, our method provides a more reasonable estimation of the occluded area by using the PBD simulation. This is another contribution of our work.

\section{Discussion and Conclusion}
In this paper, we have introduced a real-to-sim registration method to initialize and effectively register a PBD simulation to a real, live surgical scene. Several real experiments have been conducted on dVRK with detailed quantitative error analysis. 
% Our method involves both the inner structures of tissues and the surface registration using a visual perception framework \cite{SuperLi2019}.
% Our method avoids damaging the inner structures of tissues and gets rid of complex and time-consuming point tracking.
Our method provides a crucial link between volumetric PBD simulations, which is necessary in model-based control, and surface reconstructions of deformable tissue based on camera images.
% Therefore, the resulting PBD simulation combined with surface reconstruction provides a more accurate representation of the real world than either individually.

For future works, we will investigate control policies for surgical automation that use the proposed real-to-sim registration. The proposed geometrical constraints are different from traditional force models using material parameters which may result inaccuracy. More constraints can be exploited to increase realistic of simulation.
% Since the control policies are based on the fundamental models of the environment, they will generalize well to a variety of tissue manipulation tasks such as tensioning and cutting.
% Furthermore, the results in this work show effective gradient estimation with respect to the particles.
% A similar technique
Furthermore, the registration gradient can be applied to optimizing a control policy for a specific tissue manipulation task using model predictive control.

% Future works will focus on close-loop control for surgical automation using our real-to-sim registration.

% if space permits
\section*{Acknowledgement}
Many thanks to Hanpeng Jiang for experimental setup.

\clearpage
{\small
\bibliographystyle{ieeetran}
\bibliography{IEEEcitation}
}

\end{document}